\documentclass[11pt,a4paper]{article}
\usepackage[hyperref]{emnlp-ijcnlp-2019}
\usepackage{times}
\usepackage{latexsym}
\usepackage{balance}

\usepackage{array}
\usepackage{amssymb}
\usepackage{amsmath}
\usepackage{bm}
\usepackage{color}
\usepackage{multirow}
\usepackage{subfig}
\usepackage{arydshln}
\usepackage{amsfonts}
\usepackage{graphicx} 
\usepackage{graphics}
\usepackage{CJKutf8}
\usepackage{standalone}
\usepackage{caption}

\usepackage{url}

\aclfinalcopy 

\usepackage{color}
\definecolor{zptu}{RGB}{18, 141, 21}
\newcommand{\zptu}[1]{\textcolor{zptu}{zptu: #1}}

\title{Towards Better Modeling Hierarchical Structure
for Self-Attention \\with Ordered Neurons}

\author{Jie Hao\thanks{~~Work done when interning at Tencent AI Lab.}\\\normalsize Florida State University\\{\normalsize \tt haoj8711@gmail.com} \And
Xing Wang\\\normalsize Tencent AI Lab\\{\normalsize \tt  brightxwang@tencent.com} \AND
Shuming Shi\\\normalsize Tencent AI Lab\\{\normalsize \tt shumingshi@tencent.com} \And 
Jinfeng Zhang\\\normalsize Florida State University\\{\normalsize \tt jinfeng@stat.fsu.edu} \And
Zhaopeng Tu\\\normalsize Tencent AI Lab\\{\normalsize \tt zptu@tencent.com}
}

\date{}

\begin{document}
\maketitle
\begin{abstract}

Recent studies have shown that a hybrid of self-attention networks (\textsc{San}s) and recurrent neural networks (\textsc{Rnn}s) outperforms both individual architectures, while not much is known about why the hybrid models work. With the belief that modeling hierarchical structure is an essential complementary between \textsc{San}s and \textsc{Rnn}s, we propose to further enhance the strength of hybrid models with an advanced variant of \textsc{Rnn}s -- {\em Ordered Neurons} \textsc{Lstm} ~\citep[\textsc{On-Lstm},][]{Shen:2019:ICLR}, which introduces a syntax-oriented inductive bias to perform tree-like composition. Experimental results on the benchmark machine translation task show that the proposed approach outperforms both individual architectures and a standard hybrid model. Further analyses on targeted linguistic evaluation and logical inference tasks demonstrate that the proposed approach indeed benefits from a better modeling of hierarchical structure.


\end{abstract}

\section{Introduction}

Self-attention networks~\citep[\textsc{San}s,][]{Lin:2017:ICLR} have advanced the state of the art on a variety of natural language processing (NLP) tasks, such as machine translation~\cite{Vaswani:2017:NIPS}, semantic role labelling~\cite{Tan:2018:AAAI}, and language representations~\cite{Devlin:2018:arxiv}. 
However, a previous study empirically reveals that the hierarchical structure of the input sentence, which is essential for language understanding, is not well modeled by \textsc{San}s~\cite{Tran:2018:EMNLP}. 
Recently, hybrid models which combine the strengths of \textsc{San}s and recurrent neural networks (\textsc{Rnn}s) have outperformed both individual architectures on a machine translation task \cite{Chen:2018:ACL}. 
We attribute the improvement to that \textsc{Rnn}s complement \textsc{San}s on the representation limitation of hierarchical structure, which is exactly the strength of \textsc{Rnn}s~\cite{Tran:2018:EMNLP}.

Starting with this intuition, we propose to further enhance the representational power of hybrid models with an advanced \textsc{Rnn}s variant --  {\em Ordered Neurons} \textsc{Lstm}~\citep[\textsc{On-Lstm},][]{Shen:2019:ICLR}. \textsc{On-Lstm} is better at modeling hierarchical structure by introducing a syntax-oriented inductive bias, which enables \textsc{Rnn}s to perform tree-like composition by controlling the update frequency of neurons.
Specifically, we stack \textsc{San}s encoder on top of \textsc{On-Lstm} encoder ({\em cascaded encoder}). \textsc{San}s encoder is able to extract richer representations from the input augmented with structure context. To reinforce the strength of modeling hierarchical structure, we propose to simultaneously expose both types of signals by explicitly combining outputs of the \textsc{San}s and \textsc{On-Lstm} encoders.

We validate our hypothesis across a range of tasks, including machine translation, targeted linguistic evaluation, and logical inference. While machine translation is a benchmark task for deep learning models, the last two tasks focus on evaluating how much structure information is encoded in the learned representations.
Experimental results show that the proposed approach consistently improves performances in all tasks, and modeling hierarchical structure is indeed an essential complementary between \textsc{San}s and \textsc{Rnn}s.

The contributions of this paper are:
\begin{itemize}
    \item We empirically demonstrate that a better modeling of hierarchical structure is an essential strength of hybrid models over the vanilla \textsc{San}s.
    

    
    \item Our study proves that the idea of augmenting \textsc{Rnn}s with ordered neurons~\cite{Shen:2019:ICLR} produces promising improvement on machine translation, which is one potential criticism of \textsc{On-Lstm}.
\end{itemize}

\section{Approach}

Partially motivated by \newcite{wang:2016:COLING} and~\newcite{Chen:2018:ACL}, we stack a \textsc{San}s encoder on top of a \textsc{Rnn}s encoder to form a cascaded encoder. 
In the cascaded encoder, hierarchical structure modeling is enhanced in the bottom \textsc{Rnn}s encoder, based on which \textsc{San}s encoder is able to extract representations with richer hierarchical information.
Let ${\bf X} = \{{\bf x}_1, \dots, {\bf x}_N\}$ be the input sequence, the representation of the cascaded encoder is calculated by
\begin{eqnarray}
    {\bf H}_{\textsc{Rnn}s}^K &=& \textsc{Enc}_{\textsc{Rnn}s} ({\bf X}),  \\
    {\bf H}_{\textsc{San}s}^L &=& \textsc{Enc}_{\textsc{San}s} ({\bf H}_{\textsc{Rnn}s}^K), 
\end{eqnarray}
where $\textsc{Enc}_{\textsc{Rnn}s}(\cdot)$ is a $K$-layer \textsc{Rnn}s encoder that reads the input sequence, and $\textsc{Enc}_{\textsc{San}s}(\cdot)$ is a $L$-layer \textsc{San}s encoder that takes the output of \textsc{Rnn}s encoder as input.

In this work, we replace the standard \textsc{Rnn}s with recently proposed \textsc{On-Lstm} for better modeling of hierarchical structure, and directly combine the two encoder outputs to build even richer representations, as described below.

\paragraph{Modeling Hierarchical Structure with Ordered Neurons}
\textsc{On-Lstm} introduces a new syntax-oriented inductive bias -- {\em Ordered Neurons}, which enables \textsc{Lstm} models to perform tree-like composition without breaking its sequential form~\cite{Shen:2019:ICLR}. Ordered neurons enables dynamic allocation of neurons to represent different time-scale dependencies by controlling the update frequency of neurons. 
The assumption behind ordered neurons is that some neurons always update more (or less) frequently than the others, and that order is pre-determined as part of the model architecture.
Formally, \textsc{On-Lstm} introduces novel ordered neuron rules to update cell state:
\begin{eqnarray}
        w_{t} &=& \tilde{f}_{t} \circ \tilde{i}_{t}, \\
        \hat{f}_{t} &=& f_{t} \circ w_{t} + (\tilde{f}_{t}-w_{t}), \\
        \hat{i}_{t} &=& i_{t} \circ w_{t} + (\tilde{i}_{t}-w_{t}), \\ 
        c_{t} &=& \hat{f}_{t} \circ c_{t-1} + \hat{i}_{t} \circ \hat{c}_{t},
\end{eqnarray}
where forget gate $f_{t}$,  input gate $i_{t}$ and state $\hat{c}_{t}$ are same as that in the standard \textsc{Lstm}~\cite{hochreiter1997long}.
The {\em master forget gate} $\tilde{f}_{t}$ and the {\em master input gate} $\tilde{i}_{t}$ are newly introduced to control the erasing and the writing behaviors respectively. $w_{t}$ indicates the overlap, and when the overlap exists ($\exists k, w_{tk}>0$), the corresponding neurons are further controlled by the standard gates $f_{t}$ and $i_{t}$.

An ideal master gate is in binary format such as $(0, 0, 1, 1, 1)$, which splits the cell state into two continuous parts: 0-part and 1-part. 
The neurons corresponding to 0-part and 1-part are updated with more and less frequencies separately, so that the information in 0-part neurons will only keep a few time steps, while the information in 1-part neurons will last for more time steps. Since such binary gates are not differentiable, the goal turns to find the splitting point $d$ (the index of the first 1 in the ideal master gate). To this end,~\newcite{Shen:2019:ICLR} introduced a new activation function: 
\begin{eqnarray}
    \textsc{cu}(\cdot) = \textsc{cumsum}(softmax(\cdot)), 
\end{eqnarray}
where $softmax(\cdot)$ produces a probability distribution (e.g.$(0.1, 0.2, 0.4, 0.2, 0.1)$) to indicate the probability of each position being the splitting point $d$. $\textsc{cumsum}$ is the cumulative probability distribution, in which the $k$-th probability refers to the probability that $d$ falls within the first $k$ positions. The output for the above example is $(0.1, 0.3, 0.7, 0.9, 1.0)$, in which different values denotes different update frequencies. It also equals to the probability of each position's value being 1 in the ideal master gate. Since this ideal master gate is binary, $\textsc{cu}(\cdot)$ is the expectation of the ideal master gate. 

Based on this activation function, the master gates are defined as
\begin{eqnarray}
        \tilde{f}_{t} &=& \textsc{cu}_f({\bf x}_t, {\bf h}_{t-1}), \\
        \tilde{i}_{t} &=& 1-\textsc{cu}_i({\bf x}_t, {\bf h}_{t-1}),
\end{eqnarray}
where ${\bf x}_t$ is the current input and ${\bf h}_{t-1}$ is the hidden state of previous step. $\textsc{cu}_f$ and $\textsc{cu}_i$ are two individual activation functions with their own trainable parameters. 



\paragraph{Short-Cut Connection}

Inspired by previous work on exploiting deep representations~\cite{Peters:2018:NAACL,Dou:2018:EMNLP}, we propose to simultaneously expose both types of signals by explicitly combining them with a simple {\em short-cut connection}~\cite{He:2016:CVPR}.

Similar to positional encoding injection in Transformer \cite{Vaswani:2017:NIPS}, we add the output of the \textsc{On-Lstm} encoder to the output of \textsc{San}s encoder:
\begin{equation}
    \widehat{\bf H} = {\bf H}_{\textsc{On-Lstm}}^K + {\bf H}_{\textsc{San}s}^L,
    \label{eq:short-cut}
\end{equation}
where ${\bf H}_{\textsc{On-Lstm}}^K \in \mathbb{R}^{N \times d}$ is the output of \textsc{On-Lstm} encoder, and ${\bf H}_{\textsc{San}s}^L \in \mathbb{R}^{N \times d}$ is output of \textsc{San}s encoder.

\section{Experiments}


We chose  machine translation, targeted linguistic evaluation and logical inference tasks to conduct experiments in this work. The first and the second tasks evaluate and analyze models as the hierarchical structure is an inherent attribute for natural language. The third task aims to directly evaluate the effects of hierarchical structure modeling on artificial language.  



\subsection{Machine Translation}

For machine translation, we used the benchmark WMT14 English$\Rightarrow$German dataset. Sentences were encoded using byte-pair encoding (BPE) with 32K word-piece vocabulary \cite{sennrich2016neural}. We implemented the proposed approaches on top of \textsc{Transformer}~\cite{Vaswani:2017:NIPS} -- a state-of-the-art \textsc{San}s-based model on machine translation, and followed the setting in previous work~\cite{Vaswani:2017:NIPS} to train the models, and reproduced their reported results. We tested on both the \emph{Base} and \emph{Big} models which differ at hidden size (512 vs. 1024), filter size (2048 vs. 4096) and number of attention heads (8 vs. 16). All the model variants were implemented on the encoder. The implementation details are introduced in Appendix A.1. Table~\ref{tab:translation} lists the results.


\begin{table}[t]
  \centering
\scalebox{0.92}{
  \begin{tabular}{c|l|r|l}
    \bf \# & \bf Encoder Architecture    &   \bf Para.    &  \bf BLEU\\  
    \hline
      \multicolumn{4}{c}{\em Base Model} \\
    \hline
    1 & 6L \textsc{San}s     &  88M   &      27.31\\
    2 &6L \textsc{Lstm}    &  97M  &  27.23 \\
    3 &6L \textsc{On-Lstm} &  110M  &   27.44\\
    \hline
    4 &6L \textsc{Lstm} + 4L \textsc{San}s       & 104M &  27.78$^\uparrow$\\
    5 &6L \textsc{On-Lstm} + 4L \textsc{San}s    & 123M &  28.27$^\Uparrow$ \\
    6 & 3L \textsc{On-Lstm} + 3L \textsc{San}s    & 99M  &  28.21$^\Uparrow$  \\
    7 & ~~~ + Short-Cut  & 99M  & \bf 28.37$^\Uparrow$  \\
    \hline
      \multicolumn{4}{c}{\em Big Model} \\
    \hline
    8 &6L \textsc{San}s     &  264M   &  28.58\\
    9 &  Hybrid Model + Short-Cut  & 308M  & \bf 29.30$^\Uparrow$  \\
  \end{tabular}
  }
  \caption{Case-sensitive BLEU scores on the WMT14 English$\Rightarrow$German translation task. ``$\uparrow/\Uparrow$'': significant over the conventional self-attention counterpart ($p < 0.05/0.01$), tested by bootstrap resampling. ``6L \textsc{San}s'' is the state-of-the-art Transformer model. ``nL \textsc{Lstm} + mL \textsc{San}s'' denotes stacking n \textsc{Lstm} layers and m \textsc{San}s layers subsequently. ``Hybrid Model'' denotes ``3L \textsc{On-Lstm} + 3L \textsc{San}s''.}
  \label{tab:translation}
\end{table}

\paragraph{Baselines} 

(Rows 1-3) Following~\newcite{Chen:2018:ACL}, the three baselines are implemented with the same framework and optimization techniques as used in~\newcite{Vaswani:2017:NIPS}. The difference between them is that they adopt \textsc{San}s, \textsc{Lstm} and \textsc{On-Lstm} as basic building blocks respectively. As seen, the three architectures achieve similar performances for their unique representational powers.

\paragraph{Hybrid Models} 

(Rows 4-7) We first followed~\newcite{Chen:2018:ACL} to stack 6 \textsc{Rnn}s layers and 4 \textsc{San}s layers subsequently (Row 4), which consistently outperforms the individual models. This is consistent with  results reported by~\newcite{Chen:2018:ACL}. In this setting, the \textsc{On-Lstm} model significantly outperforms its \textsc{Lstm} counterpart (Row 5), and reducing the encoder depth can still maintain the performance (Row 6). We attribute these to the strength of \textsc{On-Lstm} on modeling hierarchical structure, which we believe is an essential complementarity between \textsc{San}s and \textsc{Rnn}s.
In addition, the Short-Cut connection combination strategy improves translation performances by providing richer representations (Row 7). 

\paragraph{Stronger Baseline} (Rows 8-9) We finally conducted experiments on a stronger baseline -- the \textsc{Transformer-Big} model (Row 8), which outperforms its \textsc{Transformer-Base} counterpart (Row 1) by 1.27 BLEU points. As seen, our model consistently improves performance over the stronger baseline by 0.72 BLEU points, demonstrating the effectiveness and universality of the proposed approach.

\paragraph{Assessing Encoder Strategies}

\begin{table}[t]
  \centering
\scalebox{0.92}{
  \begin{tabular}{c|l|r|l}
    \bf \# & \bf Encoder Architecture    &   \bf Para.    &  \bf BLEU\\  
    \hline
     1 & 3L \textsc{On-Lstm} $\rightarrow$ 3L \textsc{San}s    & 99M  &  28.21  \\
     \hline
     2 & 3L \textsc{San}s $\rightarrow$ 3L \textsc{On-Lstm}    & 99M  &  27.39  \\
     3 & 8L \textsc{Lstm} &  102.2M   &      27.25\\
     4 & 10L \textsc{San}s     &  100.6M   &      27.76
    \end{tabular}
  }
  \caption{Results for encoder strategies. Case-sensitive BLEU scores on the WMT14 English$\Rightarrow$German translation task. ``A $\rightarrow$  B'' denotes stacking B on the top of A. The model in Row 1 is the hybrid model in Table~\ref{tab:translation}.}
  \label{tab:hybrid}
\end{table}

We first investigate the encoder stack strategies on different stack orders. From Table~\ref{tab:hybrid}, to compare with the proposed hybrid model, we stack 3-layers \textsc{On-Lstm} on the top of 3-layers \textsc{San}s (Row 2). It performs worse than the strategy in the proposed hybrid model. The result support the viewpoint that the \textsc{San}s encoder is able to extract richer representations if the input is augmented with sequential context~\cite{Chen:2018:ACL}. 

Moreover, to dispel the doubt that whether the improvement of hybrid model comes from the increasement of parameters. We investigate the 8-layers \textsc{Lstm} and 10-layers \textsc{San}s encoders (Rows 3-4) which have more parameters compared with the proposed hybrid model. The results show that the hybrid model consistently outperforms these model variants with less parameters and the improvement should not be due to more parameters.

\subsection{Targeted Linguistic Evaluation}
\begin{table}[t]
  \centering
\scalebox{0.95}{
\begin{tabular}{c | c | c | c c c}
  \multirow{2}{*}{\bf Task}     &     {\bf\textsc{S}}	&     {\bf \textsc{O}} &	\multicolumn{3}{c}{\bf Hybrid + Short-Cut}  \\
  \cline{2-6}
    &    Final   &    Final  &   ${\bf H}_O$   &   ${\bf H}_S$    &   Final\\
  \hline
  \multicolumn{6}{c}{\em Surface Tasks} \\
  \hline 
  SeLen & \em92.71 & 90.70 & 91.94 & 89.50  & 89.86 \\
  WC  & 81.79 &  76.42 & \em90.38  &  79.10 & 80.37 \\
  \hdashline
  Avg & 87.25 & 83.56 & \bf91.16  & 84.30 &  85.12\\
  \hline 
  \multicolumn{6}{c}{\em Syntactic Tasks} \\
  \hline
  TrDep	&   44.78  & 52.58 & 51.19  & 52.55 &  \em53.28 \\
  ToCo  &  84.53   & 86.32 & 86.29  & \em87.92 &  87.89 \\
  BShif &  52.66   & \em82.68 &  81.79  & 82.05 &  81.90 \\
  \hdashline
  Avg	  &  60.66  & 73.86 & 73.09  &  74.17  &  \bf74.36 \\
  \hline
  \multicolumn{6}{c}{\em Semantic Tasks} \\
  \hline
  Tense  &   84.76  & 86.00 & 83.88  & \em86.05 &  85.91 \\
  SubN   &   85.18  & 85.44 & 85.56   &  84.59  &  \em85.81 \\
  ObjN   &   81.45  & \em86.78 & 85.72  &  85.80  &  85.38 \\
  SoMo	 &  49.87  & 49.54  &   49.23   &  49.12   &  \em49.92 \\
  CoIn   &  68.97  &72.03 & 72.06   &   72.05  &  \em72.23 \\
  \hdashline
  Avg	 &  74.05  & \bf75.96 &  75.29   &  \em75.52   &  75.85\\ 
\end{tabular}
}
  \caption{Performance on the linguistic probing tasks of evaluating linguistics embedded in the learned representations. ``{\bf\textsc{S}}'' and ``{\bf \textsc{O}}'' denote the \textsc{San} and \textsc{On-Lstm} baseline models. ``${\bf H}_O$'' and ``${\bf H}_S$'' are respectively the outputs of the \textsc{On-Lstm} encoder and the \textsc{San} encoder in the hybrid model, and ``Final'' denotes the final output exposed to decoder.}
  \label{tab:probing}
\end{table}

To gain linguistic insights into the learned representations, we conducted probing tasks~\cite{conneau:2018:acl} to evaluate linguistics knowledge embedded in the final encoding representation learned by model, as shown in Table~\ref{tab:probing}. We evaluated \textsc{San}s and proposed hybrid model with Short-Cut connection on these 10 targeted linguistic evaluation tasks. The tasks and model details are described in Appendix A.2.

Experimental results are presented in Table~\ref{tab:probing}. Several observations can be made here. The proposed hybrid model with short-cut produces more informative representation in most tasks (``Final'' in ``\textsc{S}'' vs. in ``Hybrid+Short-Cut''), indicating that the effectiveness of the model. The only exception are surface tasks, which is consistent with the conclusion in \newcite{conneau:2018:acl}: as a model captures deeper linguistic properties, it will tend to forget about these superficial features. Short-cut further improves the performance by providing richer representations (``${\bf H}_S$'' vs. ``Final'' in ``Hybrid+Short-Cut''). Especially on syntactic tasks, our proposed model surpasses the baseline more than 13 points (74.36 vs. 60.66) on average, which again verifies that \textsc{On-Lstm} enhance the strength of modeling hierarchical structure for self-attention.

\subsection{Logical Inference} 
\begin{figure}[t]
    \centering
    \includegraphics[width=0.45\textwidth]{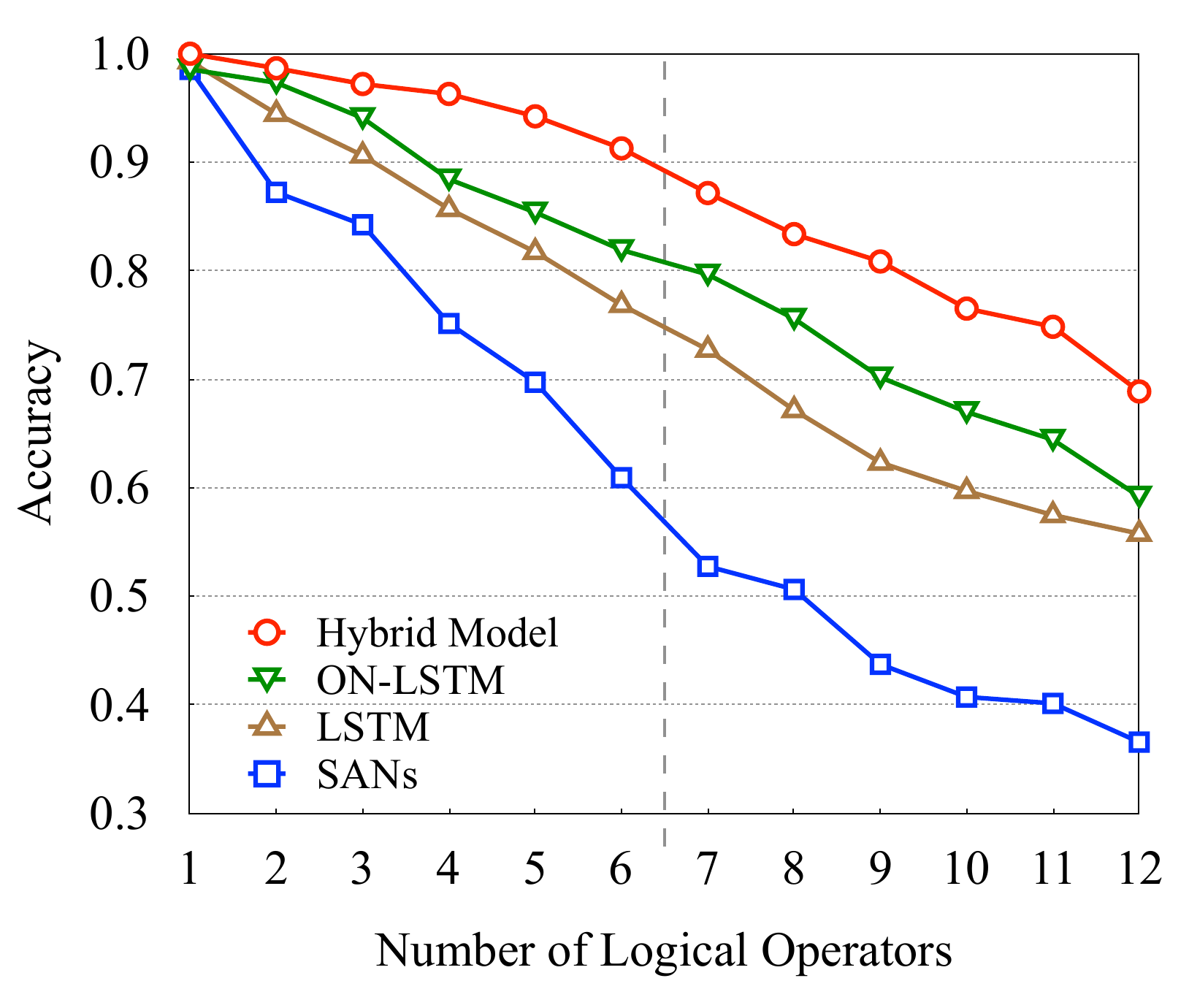}
    \caption{Accuracy of logical inference when training on logic data with at most 6 logical operators in the sequence.} 
    \label{fig:Logical_infer}
\end{figure}

We also verified the model's performance in the logical inference task proposed by ~\newcite{Bowman:2015:arXiv}. This task is well suited to evaluate the ability of modeling hierarchical structure. Models need to learn the hierarchical and nested structures of language in order to predict accurate logical relations between sentences \cite{Bowman:2015:arXiv,Tran:2018:EMNLP,Shen:2019:ICLR}. The artificial language of the task has six types of words \{a, b, c, d, e, f\} in the vocabulary and three logical operators \{or, and, not\}. The goal of the task is to predict one of seven logical relations between two given sentences. These seven relations are: two entailment types $(\sqsubset, \sqsupset)$, equivalence $(\equiv)$, exhaustive and non-exhaustive contradiction $(\land, \mid)$, and semantic independence $(\#, \smile)$. 

We evaluated the \textsc{San}s, \textsc{Lstm}, \textsc{On-Lstm} and proposed model. We followed \newcite{Tran:2018:EMNLP} to use two hidden layers with Short-Cut connection in all models. The model details and hyperparameters are described in Appendix A.3. 

Figure~\ref{fig:Logical_infer} shows the results. The proposed hybrid model outperforms both the \textsc{Lstm}-based and the \textsc{San}s-based baselines on all cases. Consistent with \newcite{Shen:2019:ICLR}, on the longer sequences ($\geq 7$) that were not included during training, the proposed model also obtains the best performance and has a larger gap compared with other models than on the shorter sequences ($\le 6$), which verifies the proposed model is better at modeling more complex hierarchical structure in sequence. It also indicates that the hybrid model has a stronger generalization ability.

\section{Related Work}


\paragraph{Improved Self-Attention Networks} 
Recently, there is a large body of work on improving \textsc{San}s in various NLP tasks~\cite{Yang:2018:EMNLP,Wu:2018:ACL,Yang:2019:AAAI,Yang:2019:NAACL,Guo:2019:AAAI,Wang:2019:ACL,sukhbaatar:2019:ACL}, as well as image classification~\cite{bello:2019:attention} and automatic speech recognition~\cite{mohamed:2019:transformers} tasks. In these works, several strategies are proposed to improve the utilize \textsc{San}s with the enhancement of local and global information. In this work, we enhance the \textsc{San}s with the \text{On-Lstm} to form a hybrid model~\cite{Chen:2018:ACL}, and thoroughly evaluate the performance on machine translation, targeted linguistic evaluation, and logical inference tasks.  

\paragraph{Structure Modeling for Neural Networks in NLP} 
Structure modeling in NLP has been studied for a long time as the natural language sentences inherently have hierarchical structures~\cite{chomsky2014aspects,bever1970cognitive}. With the emergence of deep learning, tree-based models have been proposed to integrate syntactic tree structure into Recursive Neural Networks~\cite{Socher:2013:EMNLP}, \textsc{Lstm}s~\cite{Tai:2015:ACL}, \textsc{Cnn}s~\cite{Mou:2016:AAAI}. As for \textsc{San}s,  \newcite{Hao:2019:EMNLPa},  \newcite{Ma:2019:NAACL} and \newcite{Wang:2019:EMNLP} enhance the \textsc{San}s with neural syntactic distance, multi-granularity attention scope and structural position representations, which are generated from the syntactic tree structures. 

Closely related to our work, \newcite{Hao:2019:NAACL} find that the integration of the recurrence in \textsc{San}s encoder can provide more syntactic structure features to the encoder representations. Our work follows this direction and empirically evaluates the structure modelling on the related tasks.

\section{Conclusion}

In this paper, we adopt the \textsc{On-Lstm}, which models tree structure with a novel activation function and structured gating mechanism, as the \textsc{Rnn}s counterpart to boost the hybrid model. We also propose a modification of the cascaded encoder by explicitly combining the outputs of individual components, to enhance the ability of hierarchical structure modeling in a hybrid model. Experimental results on machine translation, targeted linguistic evaluation and logical inference tasks show that the proposed models achieve better performances by modeling hierarchical structure of sequence. 

\section*{Acknowledgments}
J.Z. was supported by the National Institute of General Medical Sciences of the National Institute of Health under award number R01GM126558. 
We thank the anonymous reviewers for their insightful comments.

\bibliography{emnlp-ijcnlp-2019}
\bibliographystyle{acl_natbib}

\appendix

\section{Supplemental Material}

\subsection{Machine Translation}
We conducted experiments on the widely-used WMT14 English$\Rightarrow$German dataset\footnote{http://www.statmt.org/wmt14/translation-task.html} consisting of about 4.56M sentence pairs. We used newstest2013 and newstest2014 as development set and test set respectively. We applied byte pair encoding (BPE) toolkit\footnote{https://github.com/rsennrich/subword-nmt} with 32K merge operations. The case-sensitive NIST BLEU score~\cite{papineni2002bleu} is used as the evaluation metric. All models were trained on eight NVIDIA Tesla P40 GPUs where each was allocated with a batch size of 4096 tokens.

For \emph{Base} model, it has embedding size and hidden size of 512, filter size of 2048 and attention heads of 8. Compared with \emph{Base} model, \emph{Big} model has embedding size and hidden size of 1024, filter size of 4096 and attention heads of 16.
For both \emph{Base} and \emph{Big} models, the number of encoder and decoder layer is 6, all types of dropout rate is 0.1. Adam~\cite{kingma2015adam} is used with $\beta_1 = 0.9$, $\beta_2 = 0.98$ and $\epsilon = 10^{-9}$. The learning rate is 1.0 and linearly warms up over the first 4,000 steps, then decreases proportionally to the inverse square root of the step number. Label smoothing is 0.1 during training \cite{DBLP:conf/cvpr/SzegedyVISW16}. All the results we reported are based on the individual models without using the averaging model or ensemble.

\subsection{Targeted Linguistic Evaluation}
We conducted 10 probing tasks\footnote{https://github.com/facebookresearch/SentEval/tree/master\\/data/probing} to study what linguistic properties are captured by the encoders~\cite{conneau:2018:acl}.
A probing task is a classification problem that focuses on simple linguistic properties of sentences. `SeLen' predicts the length of sentences in terms of number of words. `WC' tests whether it is possible to recover information about the original words given its sentence embedding. `TrDep' checks whether an encoder infers the hierarchical structure of sentences. In `ToCo' task, sentences should be classified in terms of the sequence of top constituents immediately below the sentence node. `BShif' tests whether two consecutive tokens within the sentence have been inverted. `Tense' asks for the tense of the main-clause verb. `SubN' focuses on the number of the main clause's subject. `ObjN' tests for the number of the direct object of the main clause. In `SoMo', some sentences are modified by replacing a random noun or verb with another one and the classifier should tell whether a sentence has been modified. `CoIn' contains sentences made of two coordinate clauses. Half of sentences are inverted the order of the clauses and the task is to tell whether a sentence is intact or modified.

Each of our probing model consists a pre-trained encoder of model variations from machine translation followed by a MLP classifier~\cite{conneau:2018:acl}. The mean of the encoding layer is served as the sentence representation passed to the classifier. The MLP classifier has a dropout rate of 0.3, a learning rate of 0.0005 with Adam optimizer and were trained for 250 epochs. 

\subsection{Logical Inference}
We used the artificial data\footnote{https://github.com/sleepinyourhat/vector-entailment} described in ~\newcite{Bowman:2015:arXiv}. The train/dev/test dataset ratios are set to 0.8/0.1/0.1 with the number of logical operations range from 1 to 12.
We followed ~\newcite{Tran:2018:EMNLP} to implement the architectures: premise and hypothesis sentences are encoded in fixed-size vectors, which are concatenated and fed to a three layer feed-forward network for classification of the logical relation. For \textsc{Lstm} based models, we took the last hidden state of the top layer as a fixed-size vector representation of the sentence. For the hybrid and \textsc{San}s models, we used two trainable queries to obtain the fixed-size representation.  

In our experiments, both word embedding size and hidden size are set to 256. All models have two layers, a dropout rate of 0.2, a learning rate of 0.0001 with Adam optimizer, and were trained for 100 epochs. Especially, for hybrid model, we stacked one \textsc{On-Lstm} layer and one \textsc{San}s layer subsequently. Short-Cut connection between layers is added into all models for fair comparison. 

\end{document}